\documentclass[letterpaper, 10 pt, conference]{ieeeconf}

\IEEEoverridecommandlockouts
\usepackage[english]{babel}
\usepackage[protrusion=false,stretch=10,shrink=10]{microtype}
\PassOptionsToPackage{hyphens}{url}
\ifx\pdfoutput\undefined
    \usepackage[hypertex]{hyperref}
\else
    \usepackage[pdftex,hypertexnames=false]{hyperref}
\fi
\usepackage{cite}
\usepackage{amsmath, amssymb, amsfonts}
\usepackage{graphicx}
\usepackage{dblfloatfix}   
\usepackage{caption}
\usepackage{subcaption}
\usepackage{siunitx}
\let\labelindent\relax % to prevent an error
\usepackage{enumitem}
\usepackage{listings}
\usepackage{algorithmic}
\usepackage{booktabs}
\usepackage{textcomp}

\usepackage{xcolor}
\definecolor{myindigo}{HTML}{332288}
\definecolor{mycyan}{HTML}{88CCEE}
\definecolor{myteal}{HTML}{44AA99}
\definecolor{mygreen}{HTML}{117733}
\definecolor{myolive}{HTML}{999933}
\definecolor{mysand}{HTML}{DDCC77}
\definecolor{myrose}{HTML}{CC6677}
\definecolor{mywine}{HTML}{882255}
\definecolor{mypurple}{HTML}{AA4499}
\definecolor{mygray}{HTML}{DDDDDD}

\begin{document}

% \title{Sound and Consequences: Should All Robots Sound Cute?}
\title{Exploring Consequential Robot Sound: \\Should We Make Robots Quiet and Kawaii-et?}

\author{Brian J. Zhang$^1$, Knut Peterson$^1$, Christopher A. Sanchez$^2$, Naomi T. Fitter$^1$% <-this % stops a space
\thanks{$^1$ Collaborative Robotics and Intelligent Systems (CoRIS) Institute, Oregon State University, Corvallis, OR 97331, USA.
        {\tt\footnotesize \{zhangbr, fittern\}@oregonstate.edu};
        {\tt\footnotesize petersk@rose-hulman.edu}}
\thanks{$^2$ School of Psychological Science, Oregon State University, Corvallis, OR 97331, USA.
        {\tt\footnotesize sancchri@oregonstate.edu}}
}

\maketitle

\begin{abstract}
All robots create consequential sound---sound produced as a result of the robot's mechanisms---yet little work has explored how sound impacts human-robot interaction. Recent work shows that the sound of different robot mechanisms affects perceived competence, trust, human-likeness, and discomfort. However, the physical sound characteristics responsible for these perceptions have not been clearly identified. In this paper, we aim to explore key characteristics of robot sound that might influence perceptions. A pilot study from our past work showed that quieter and higher-pitched robots may be perceived as more competent and less discomforting. To better understand how variance in these attributes affects perception, we performed audio manipulations on two sets of industrial robot arm videos within a series of four new studies presented in this paper. Results confirmed that quieter robots were perceived as less discomforting. In addition, higher-pitched robots were perceived as more energetic, happy, warm, and competent. Despite the robot's industrial purpose and appearance, participants seemed to prefer more ``cute'' (or ``kawaii'') sound profiles, which could have implications for the design of more acceptable and fulfilling sound profiles for human-robot interactions with practical collaborative robots.
\end{abstract}

\section{Introduction}

Like all machines, robots invariably produce sound alongside their motion in the form of \emph{consequential sound}~\cite{langeveld_product_2013}. Robot sound can affect human-robot interaction (HRI) by reducing the perceived usability of prosthetics~\cite{castellini_upper-limb_2016}, improving human-robot localization~\cite{cha_effects_2018}, or even impeding interactions with robot companions~\cite{inoue_effective_2008}. In the most extreme case, excessively loud sound has precluded the usage of a robot~\cite{seck_marine_2015}. Despite these examples, however, little is known about robot consequential sound and how it might impact successful design for HRI. This paper aims to \emph{identify important physical characteristics of robot consequential sound} and measure the strength and consistency of these characteristics' effects on the \emph{perception of robot arm recordings}.

\begin{figure}[t]
    \centering
    \includegraphics[width = 3.2in]{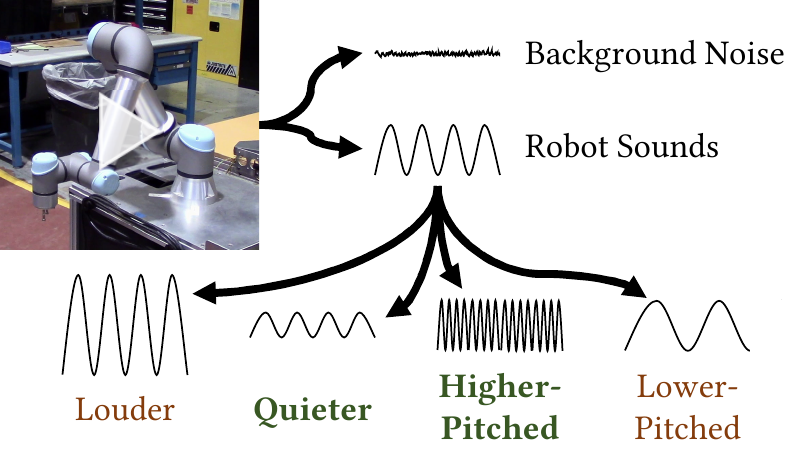}
    \caption{An overview of the study manipulations.}
    \vspace{-0.1in}
    \label{fig:teaser}
\end{figure}

Past researchers have encountered considerable challenges when attempting to connect objective auditory characteristics to the subjective perception of robot consequential sounds. A survey of servo motor sounds found that the correlations between the spectral centroid, servo weight, and subjective audio strength were the only significant relationships between twelve objective measures and six subjective measures~\cite{moore_making_2017}. Conversely, a survey with high-end and low-end robot arm sounds failed to find a significant difference between the two~\cite{tennent_good_2017}. These past efforts relied only on sound samples from existing robot arms and mechanisms (rather than richer audio-edited sets of existing and prospective robot sounds), which limited the work's ability to acutely understand how specific characteristics of sound impacted perception. Other research has considered audio-edited versions of consequential robot sound. Work in~\cite{cha_effects_2018} showed that adding broadband and tonal sound to robot movement makes a robot more localizable and noticeable, though also more annoying. Recent work by the authors has pointed towards the benefit of quieter and higher-pitched robots, though a lack of granularity in conditions limits the understanding of this effect~\cite{zhang_consequential_2021}. 
While these past results demonstrate how altering the sound profile of robots can improve experiences working with collaborative robots, \emph{key questions remain as to how finer audio manipulations on promising objective scales, such as those shown in Fig.~\ref{fig:teaser}, will influence subjective robot perception}. 

To understand the current state of robot sound research, we reviewed related work in Section~\ref{sec:relatedwork}. 
Based on our previous pilot study results in~\cite{zhang_consequential_2021}, we designed and conducted follow-up studies as described in Section~\ref{sec:methods}. The results of \emph{manipulations in loudness} are presented in Section~\ref{sec:loudness}, and the results of \emph{manipulations in pitch} are presented in Section~\ref{sec:pitch}. Section~\ref{sec:discussion} discusses our findings and related design implications for robot sound profiles. 

\section{Related Work}
\label{sec:relatedwork}

To inform our efforts, we explored prior work in robot sound and related fields; the nascent field of robot sound can benefit from drawing on more established bodies of work. 

\vspace{0.07in}
\noindent\emph{Psychoacoustics \& Sound Design:}
Psychoacoustics defines the three subjective qualities of sound as \emph{loudness}, \emph{pitch}, and \emph{timbre}. While \emph{loudness} and \emph{pitch} are well-known to non-psychoacousticians, the \emph{timbre} of a sound, which can be broken down into factors such as sharpness, pleasantness, roughness, and tonalness, encompasses all remaining perceptual properties of a sound~\cite{fastl_psychoacoustics_2007}. However, sound design for products and interactions must also take context-dependent factors (e.g., the input of other senses, personal knowledge and associations of sound) into account ~\cite{blauert_sound-quality_1997}. For instance, modulation in mechanical sounds (e.g., gearboxes) is generally undesirable while modulation in musical sounds (e.g., vocoders) is often desirable~\cite{lyon_product_2000}; loudness may be perceived as unwanted noise, but products such as vacuums and motorcycles may benefit from increased loudness~\cite{lyon_designing_2000}. As such, good sound design usually relies on a combination of proper sound design methods and expertise in related fields such as acoustics, engineering, and psychology~\cite{lyon_product_2000, langeveld_product_2013, ozcan_product_2008}. 

Product sound design defines the two main categories of product sound as \emph{consequential sound} (e.g., the meshing of servo motor gears, the hum of cooling fans, or the rattle of drivetrain chains) and \emph{intentional sound} (e.g., non-linguistic utterances, back-up alarm beeps, or played-back music). As their origins and design spaces differ greatly, consequential and intentional sound require different design processes and expertise~\cite{langeveld_product_2013}. Characterizing and designing consequential sound, which relies on product components' designs, materials, and manufacturing, is especially difficult~\cite{langeveld_product_2013, blauert_sound-quality_1997}. As a first step in this challenging process, we aim to \emph{identify key perceptual trends related to loudness and pitch of consequential robot sounds.}

\vspace{0.07in}
\noindent\emph{Sound in Human-Robot Interaction:}
Sound in HRI also falls into intentional and consequential sound categories. Recent work has explored consequential robot sounds and subtypes of intentional robot sounds such as \emph{vocable} and \emph{transformative} sound. Vocable sound, or non-linguistic utterances, can independently convey affect and amplify the affective interpretation of actions~\cite{read_situational_2014}. Transformative sound, or intentional sound designed to be associated with robot motion, also changes affective and objective responses. In a study on localization of a visually obscured robot, playing broadband and tonal sounds increased the accuracy and inference speed for localizing the robot as well as perceived noticeability, localizability, and annoyance~\cite{cha_effects_2018}. Work on masking undesirable consequential sound with musical sounds showed that participants rated masked sound more positively than just consequential sound~\cite{trovato_sound_2018, zhang_bringing_2021}. 

Consequential robot sound research has evaluated sound at the component, robot, and inter-robot level. For robot components, a study on servomotor sounds found that although the sounds were separable on six subjective attribute scales, only one of these attributes was correlated with objective characteristics of the servomotors~\cite{moore_making_2017}. 
Yet, participants were able to accurately identify the lowest-quality and most inexpensive servomotor as the most inappropriate, untrustworthy, weak, imprecise, and inexpensive~\cite{moore_unintended_2019}.
Another project found that consequential sounds of a humanoid robot may communicate emotions separately from the corresponding robot motions, potentially requiring masking through sonification of motion~\cite{frid_perception_2018}. 
Lastly, comparisons between the sounds of a high-end and a low-end robot arm found that the effect of sound trended differently across contexts, but the presence of sound detracted overall~\cite{tennent_good_2017}. While these investigations do compare current robot sounds, they unfortunately do not offer clear direction on how to design \emph{future} robot sound. Our work builds upon this body of research by \emph{investigating how systematic  manipulations of consequential robot sound might affect perceptions of a robot.}

\section{Methods: All Studies}
\label{sec:methods}

We conducted a series of four studies via Amazon Mechanical Turk (MTurk) to explore the effects of modifying loudness and pitch on participant perceptions of robots. All study procedures were approved by Oregon State University under IRB-2019-0172. 

\subsection{Hypotheses}

These hypotheses arose from our past results in~\cite{zhang_consequential_2021}:

\begin{enumerate}[label=\textbf{H{\arabic*}:}, leftmargin=.5in]
    \item Reducing the loudness of a robot will lead to perceptions of lower discomfort and higher competence. 
    
    \item Lowering the pitch of a robot will lead to perceptions of higher discomfort.
\end{enumerate}

\subsection{Study Design}

\begin{figure*}[t]
    \centering
    \vspace{0.055in}
    \includegraphics[width=5.7in]{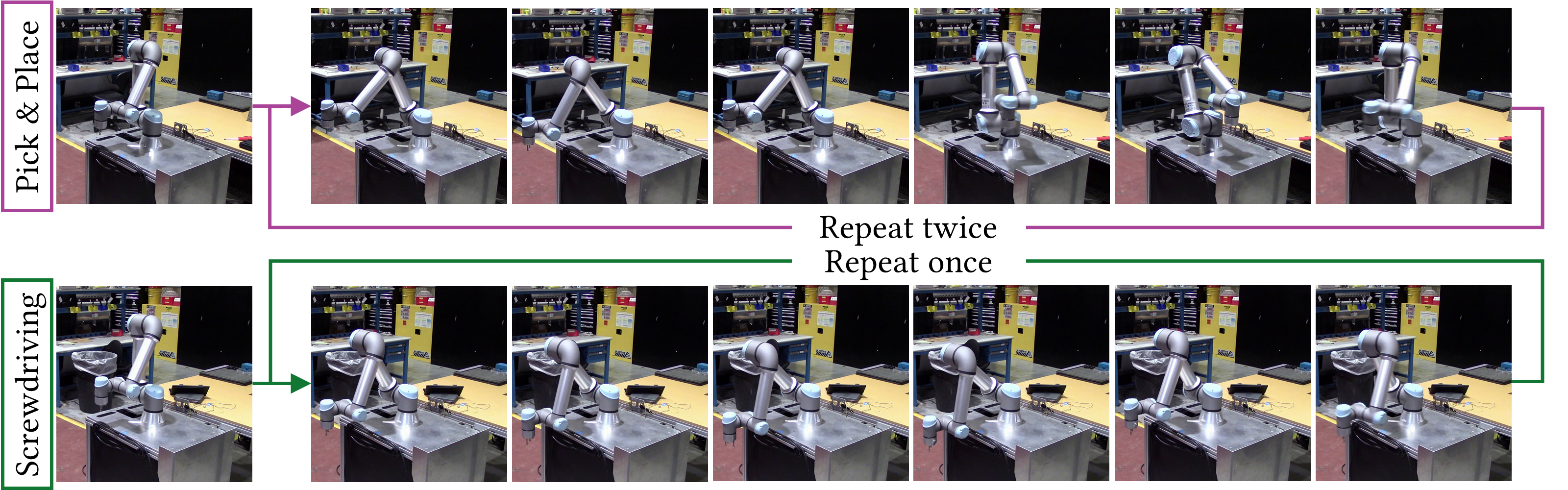}
    \vspace{-0.1in}
    \caption{Cropped keyframes from the screwdriving and pick \& place motions.}
    \vspace{-0.15in}
    \label{fig:ur5e-snapshots}
\end{figure*}

The studies employed a UR5e robot arm. The arm performed motions mimicking \emph{pick \& place} and \emph{screwdriving} tasks without any props, as shown in Fig.~\ref{fig:ur5e-snapshots}. We performed manipulations across two video types to serve as repetition for confirming trends across different robot use contexts.

Using a Canon Vixia HF R800 camera and a Blue Snowball microphone, the arm was recorded completing these motions with different background objects. After we captured the recordings, we manipulated the audio track of the videos to create different stimuli. All stimuli included noise reduction with a level of 20~dB, 3.50 sensitivity, and 2 frequency smoothing bands and then underwent condition-specific manipulation before background noise was added back into the stimuli. Manipulations in loudness were performed in MATLAB and manipulations in pitch were performed in Audacity using the Change Pitch tool with the high-quality setting. The \emph{pick \& place} motion stimuli were 18 seconds long and the \emph{screwdriving} motion stimuli were 14 seconds long. The accompanying video compares several select stimuli, while  the full set of stimuli are available in~\cite{zhang_appendix_2021}.

\subsection{Participants}

Participants were MTurk workers from the United States with qualifications of $>$97\% prior task approval rate and $>$5000 previously approved tasks. Study instructions required that participants have normal or corrected-to-normal vision and hearing and also be native English speakers. Participants were not allowed to participate more than once across all studies. We compensated participants with 3.75 USD for finishing the estimated 15-minute task. 
Based on the results from~\cite{zhang_consequential_2021}, we performed an \emph{a priori} power analysis using G*Power for a repeated-measures ANOVA with $\alpha=0.05$ and $\beta=0.05$~\cite{erdfelder_gpower_1996}. As the effect sizes ranged from small to medium, a sample of $N=100$ participants was recommended for each study to ensure adequate statistical power.

\subsection{Procedure}

Upon enrolling in the study, participants first were asked to provide informed consent. Participants then completed a 15-minute survey implemented via Qualtrics, beginning with an introductory module common across all studies. This module included a loudness calibration video, which requested that the respondent wear headphones or earbuds and set their device volume so that the speech in the video was at a ``loud but comfortable volume.'' The survey also requested that the participant not change their device volume for the remainder of the study. Next, the participant was introduced to the UR5e via an image of the UR5e in the lab environment and the following explanation: \emph{``We would like to know what you think about several versions of a robot. Here is a photo of the robot in a lab environment.''} Lastly, the module presented a video stimulus featuring the Cassie robot walking on a treadmill and a post-stimulus questionnaire as described in Section~\ref{sec:methods-measurement}. We included this introductory stimulus across all studies to have a common point of reference for all participants.

After finishing the introductory module, participants completed the study-specific module, which included eight levels of manipulation for either \emph{loudness} or \emph{pitch} on either the \emph{pick \& place} or \emph{screwdriving} stimuli. Studying eight levels of manipulation is consistent with prior work in the robot sound field~\cite{thiessen_infrasound_2019, moore_making_2017}. The order of stimuli was counterbalanced such that stimuli of adjacent levels would not play directly after one another, and the order of stimuli was presented as evenly as possible. For each stimulus, the survey page hid the ``Next'' button for 40 seconds and requested the participant watch the stimulus video at the top of the page. The participant was also asked to answer the post-stimulus questionnaire as described in Section~\ref{sec:methods-measurement}. After the fourth and eighth stimulus presentations, participants had to successfully complete an attention check question to continue. This step was to help ensure that participants were attending to the task and not selecting random responses.

Lastly, participants completed the closing module. This module contained a free-response question that asked participants 
to describe the parts of the robot videos that affected their responses the most in a minimum of 200 characters.
Next, the survey contained a manipulation check to determine whether participants could differentiate between stimuli when shown the videos side-by-side. Participants who reported that the stimuli were the same did not advance. Participants who reported that the stimuli were different then completed the Negative Attitudes Towards Robots Scale (NARS) and a demographic survey, both described in Section~\ref{sec:methods-measurement}, before receiving a completion code for compensation. 

Participants who did not finish the survey, failed attention or manipulation checks, or attempted to take the survey more than once were excluded from the study.

\subsection{Measurement}
\label{sec:methods-measurement}

We recorded these subjective and perceptual measures:

\begin{itemize}
    \item \emph{Post-stimulus questionnaire}: after each stimulus, the Robotic Social Attributes Scale (RoSAS) captured participant perceptions of \emph{warmth}, \emph{competence}, and \emph{discomfort} by combining six component attributes for each subscale~\cite{carpinella_robotic_2017}. Participants rated each attribute on a six-point Likert scale from ``definitely not associated'' to ``definitely associated.'' \emph{Valence} and \emph{arousal} from the circumplex model of affect were acquired via participant association of the robot with ``happy,'' which also contributes to the \emph{warmth} subscale, and an additional attribute of ``energetic.'' After all stimuli, a free-response question requested the most important factors in participant responses. 
    
    \item \emph{Attitudes questionnaire}: after the post-stimulus questionnaires, the NARS questions captured general attitudes towards robots for potential use in subsequent covariate analyses. Using a seven-point Likert scale, participants indicated their agreement level with fourteen statements that were combined into subscales of negative attitudes towards \emph{interactions with robots}, \emph{social influence of robots}, and \emph{emotions in robots} as described in~\cite{syrdal_negative_2009}.
    
    \item \emph{Demographic questionnaire}: at the end of the survey, a final questionnaire recorded demographic and occupational information.
\end{itemize}
This work focuses on the analysis of post-stimulus responses. The questionnaires and stimulus videos are available in~\cite{zhang_appendix_2021}.

\begin{figure*}
    \centering
    \vspace{0.055in}
    \includegraphics[width=6.31in]{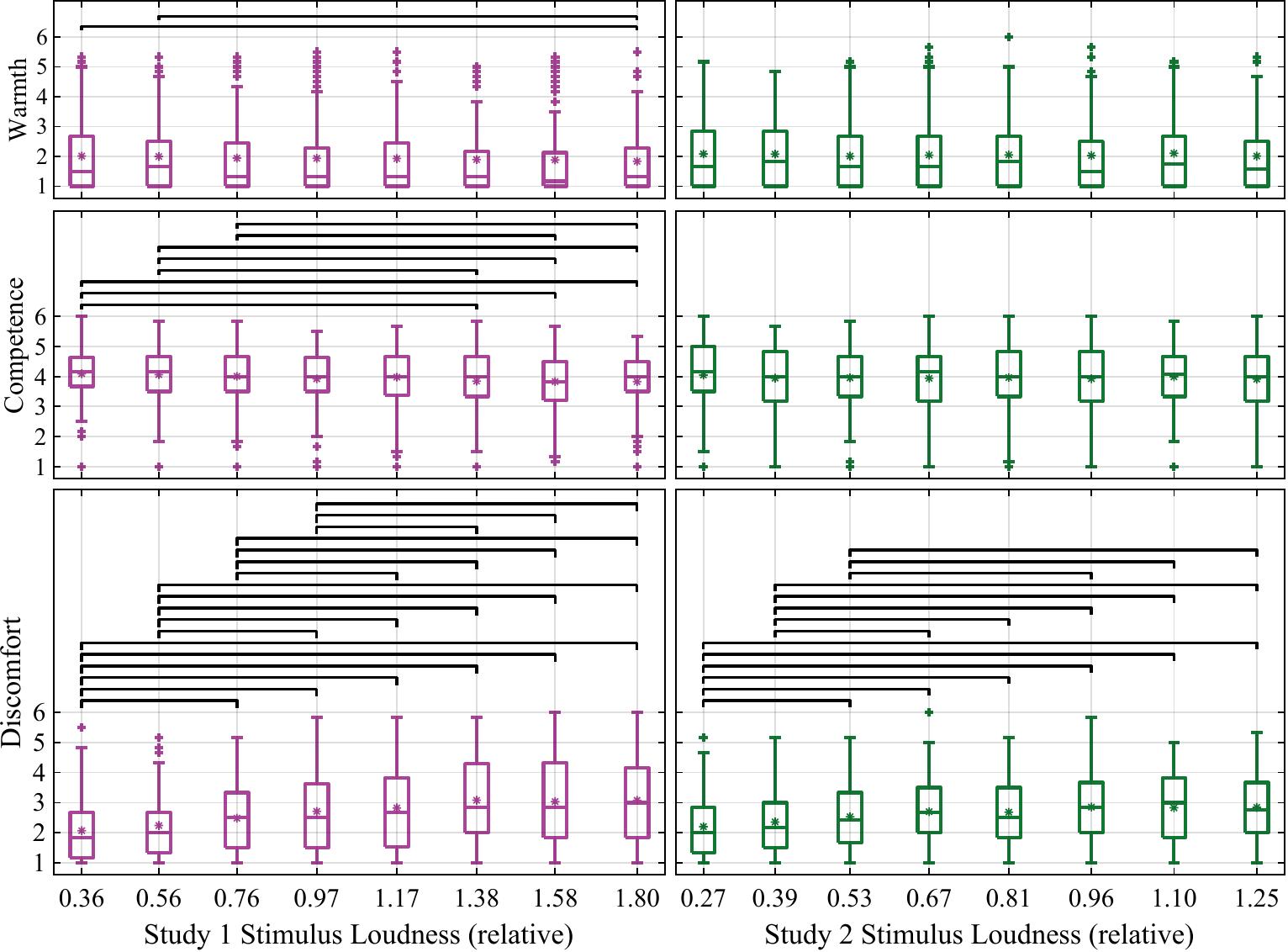}
    \caption{Post-stimulus RoSAS responses for Studies 1-2. Boxplots include boxes from the 25th to the 75th percentiles, center lines for medians, asterisks for means, whiskers up to 1.5 times the inter-quartile range, and ``+'' marks for outliers. Brackets above the boxplots indicate pairwise differences.}
    \vspace{-0.1in}
    \label{fig:rosas-loud}
\end{figure*}

\subsection{Analysis}

Responses to the post-stimulus questionnaire were analyzed using repeated-measures analysis of variance (rANOVA) tests with an $\alpha = 0.05$ significance level and adjusted with Greenhouse-Geisser sphericity corrections. 
Pairwise comparisons were conducted to unpack any significant main effects using Tukey's HSD test, which adjusts for Type 1 error inflation.
We report effect size $\eta^2$, where $\eta^2 = 0.010$ is considered a small effect, $\eta^2 = 0.040$ a medium effect, and $\eta^2 = 0.090$ a large effect~\cite{funder_evaluating_2019}. All statistical analyses were conducted using jamovi~\cite{the_jamovi_project_jamovi_2020, r_core_team_r_2020, singmann_afex_2018, lenth_emmeans_2020}.

\section{Results: Studies 1-2, on Loudness}
\label{sec:loudness}

These studies tested \textbf{H1} to determine how loudness might impact perception of robots. Study~1 included \emph{pick \& place} stimuli with 35.5\% to 179.6\% of the N5 loudness compared to the loudness calibration video~\cite{fastl_psychoacoustics_2007}. Study~2 included \emph{screwdriving} stimuli with 26.7\% to 124.9\% relative loudness. 

\subsection{Participants}

Study~1 was completed by $N=99$ participants who were adults between 21 and 69 years of age ($M=36.3$, $SD=10.9$), with 62.6\% cisgender men, 35.4\% cisgender women, 1.0\% transgender women, and 1.0\% non-binary individuals. An additional 14 respondents failed the manipulation check, and their responses are not considered in these results.

Study~2 was completed by $N=94$ participants who were adults between 22 and 69 years of age ($M=38.9$, $SD=11.7$), with 54.3\% cisgender men, 43.6\% cisgender women, and 2.1\% non-binary individuals. An additional 12 respondents failed the manipulation check for this study and were excluded from the results.

\begin{figure*}
    \centering
    \vspace{0.055in}
    \includegraphics[width=6.31in]{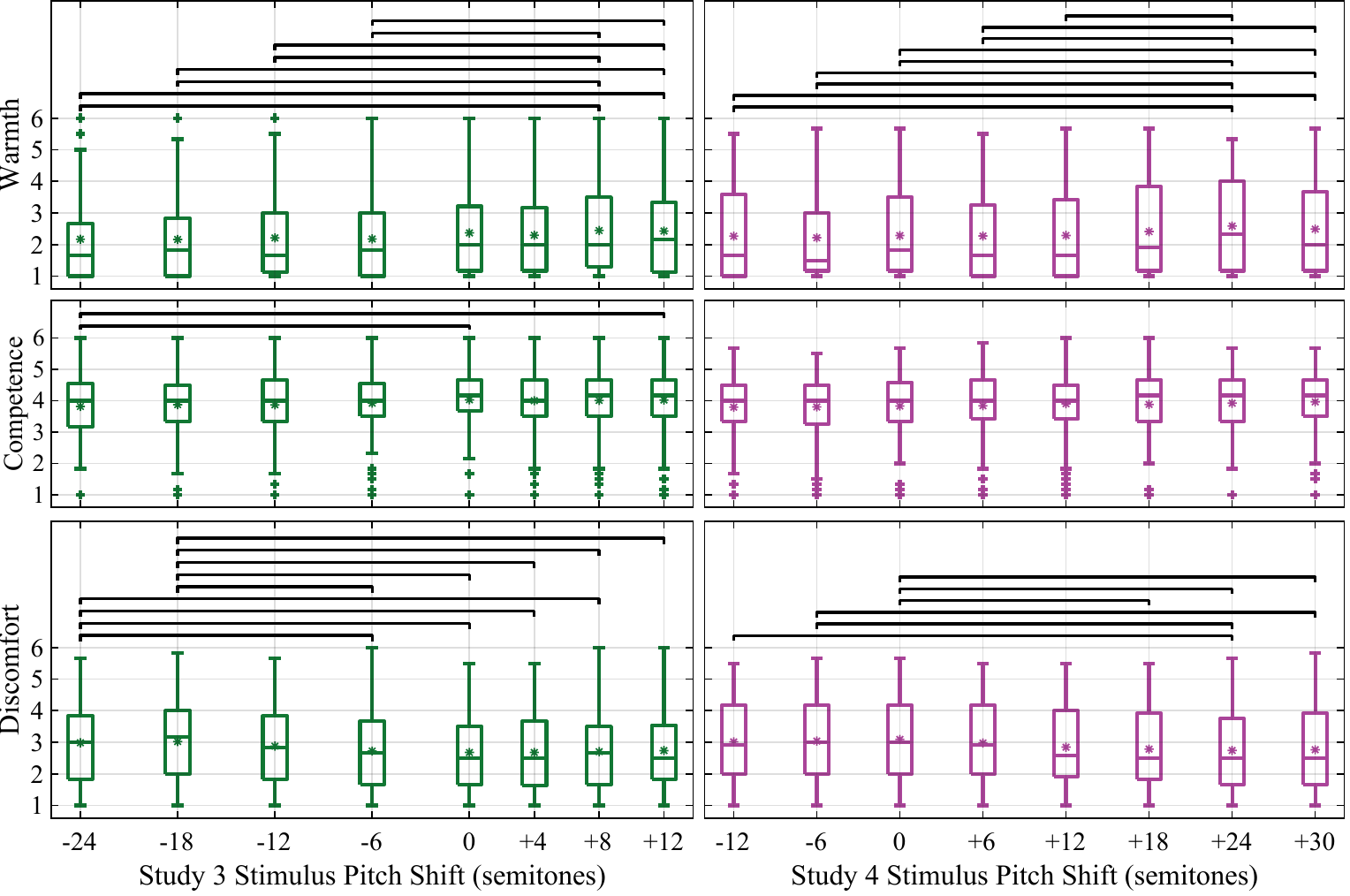}
    \caption{Post-stimulus RoSAS responses for Studies 3-4. Boxplots include boxes from the 25th to the 75th percentiles, center lines for medians, asterisks for means, whiskers up to 1.5 times the inter-quartile range, and ``+'' marks for outliers. Brackets above the boxplots indicate pairwise differences.}
    \vspace{-0.1in}
    \label{fig:rosas-pitch}
\end{figure*}

Introductory stimulus and NARS results did not differ significantly between Study~1 and Study~2, suggesting that participants across these studies were well-matched on these initial robot attitudes.

\subsection{Effects of Loudness}

Responses to post-stimulus questionnaires along RoSAS subscales for Studies 1-2 appear in Fig.~\ref{fig:rosas-loud}. rANOVAs showed that, as \emph{loudness} increased: perceived \emph{valence} decreased in Study~1 ($p<0.001$, $F(5.26,\,515.82)=4.16$, $\eta^2=0.007$) but not in Study~2; perceived \emph{arousal} increased in Study~2 ($p=0.003$, $F(5.74,\,533.80)=3.38$, $\eta^2=0.008$) but not in Study~1;  perceived \emph{warmth} decreased in Study~1 ($p=0.011$, $F(5.65,\,553.17)=2.88$, $\eta^2=0.029$) but not in Study~2; perceived \emph{competence} decreased in Study~1 ($p<0.001$, $F(5.67,\,555.56)=6.86$, $\eta^2=0.011$) but not in Study~2; and perceived \emph{discomfort} increased in both Study~1 ($p<0.001$, $F(3.45,\,338.33)=37.74$, $\eta^2=0.077$) and Study~2 ($p<0.001$, $F(4.48,\,416.48)=16.89$, $\eta^2=0.040$).

\section{Results: Studies 3-4, on Pitch}
\label{sec:pitch}

These studies tested \textbf{H2} to determine the effect of pitch on perception of robots. Study~3 included \emph{screwdriving} stimuli frequency shifted by -24, -18, -12, -6, 0, +4, +8, +12 semitones. Study~4 included \emph{pick \& place} stimuli shifted by -12, -6, 0, +6, +12, +18, +24, +30 semitones to investigate whether trends from Study~3 persisted at higher pitches. 

\subsection{Participants}

Study~3 was completed by $N=89$ participants who were adults between 22 and 69 years of age ($M=38.5$, $SD=11.8$), with 58.4\% cisgender men and 41.6\% cisgender women. An additional 15 respondents failed the manipulation check  and were excluded from the results.

Study~4 was completed by $N=100$ participants who were adults between 23 and 69 years of age ($M=36.8$, $SD=11.1$), with 53.0\% cisgender men, 45.0\% cisgender women, and 1.0\% transgender men. An additional 6 respondents failed the manipulation check, and their responses are not considered in these results.

Introductory stimulus and NARS results did not differ significantly between Study~3 and Study~4 with the exception of perceived discomfort from the introductory stimulus ($t(187)=2.02$, $p=.04$). Thus, participants appear generally consistent on their base attitudes towards robots. 

\subsection{Effects of Pitch}

Responses to post-stimulus questionnaires along RoSAS subscales for Studies 3-4 appear in Fig.~\ref{fig:rosas-pitch}. 
rANOVAs showed that, as \emph{pitch} increased: perceived \emph{valence} increased in both Study~3 ($p<0.001$, $F(5.13,\,451.67)=5.01$, $\eta^2=0.013$) and Study~4 ($p<0.001$, $F(4.16,\,412.10)=10.32$, $\eta^2=0.021$); perceived \emph{arousal} increased in both Study~3 ($p<0.001$, $F(5.89,\,525.08)=6.69$, $\eta^2=0.022$) and Study~4 ($p<0.001$, $F(6.11,\,605.16)=4.62$, $\eta^2=0.013$);  perceived \emph{warmth} increased in both Study~3 ($p<0.001$, $F(5.09,\,448.24)=5.73$, $\eta^2=0.007$) and Study~4 ($p<0.001$, $F(4.23,\,418.67)=7.83$, $\eta^2=0.008$); perceived \emph{competence} increased in Study~3 ($p=0.004$, $F(5.59,\,491.88)=3.28$, $\eta^2=0.006$) but not in Study~4; and perceived \emph{discomfort} decreased in both Study~3 ($p<0.001$, $F(5.49,\,482.94)=5.44$, $\eta^2=0.011$) and Study~4 ($p<0.001$, $F(4.81,\,476.44)=5.00$, $\eta^2=0.011$).

\section{Discussion}
\label{sec:discussion}

Results partially supported the affective responses expected in \textbf{H1}. In Study 1, more quiet stimuli led to a lower perception of discomfort and a higher perception of competence as expected, as well as a higher perception of warmth, but perceptions of competence and warmth did not change in Study 2. Notably, Study 1 led to a significant ($p<0.001$) increase in perceived competence for more quiet stimuli with a small to medium effect size, but Study 2 did not indicate any difference on this scale. Increases in valence and warmth for quieter stimuli in Study 1 and increases in arousal for louder stimuli in Study 2 had less than small effect sizes, while decreases for discomfort for quieter stimuli had medium to large effect sizes in both studies. Generally, an \emph{increase in discomfort as the loudness increased} was the strongest and most repeatable result related to \textbf{H1}.

Data from the free-response survey field can help to explain and support the observed differences in loudness perception. One participant remarked that \emph{``when I think of robots I think of soft whirls, maybe some slight clunking,''} and objected to anything \emph{``overly noisy or off-putting.''} On a related note, other respondents noted that the robot seemed to be making \emph{``growling or roaring animal-like noise,'' ``a ferocious growl,''} or sound \emph{``like a monster or dragon,''} although the audio was sampled from the natural consequential sound of the robot. Participants labeled quieter sounds as \emph{``gentle,'' ``more safe,''} or \emph{``less dangerous.''} Conversely, louder sounds were \emph{``scary,'' ``disturbing,'' ``horrifying,'' ``dangerous,'' ``mean,'' ``aggressive,''} and \emph{``harsh.'' }
The uniform overall perception seems to be that quieter robot arms are preferable. 

Results fully supported the affective responses expected in \textbf{H2} as well as additional affective responses not anticipated. In Studies 3 and 4, \emph{higher-pitched stimuli led to higher arousal, valence, and warmth, as well as lower discomfort}. In Study 3, higher-pitched stimuli also led to higher perception of competence, though with less than small effect size. For arousal, valence, and discomfort, effect sizes ranged from small to medium, while the effect size for warmth was less than small in both studies. 

Again, free-response data can reinforce our understanding of these trends. One participant remarked that \emph{``the sounds they made changed the robot's `personality,'\thinspace''} going on to state that \emph{``some sounded cute,'' ``some sounded horrifying,''} and \emph{``a few just sounded machine-like.''} Despite the results from Studies 3 and 4, we suspect that at some point, the high pitch of a robot may ultimately be too high. Although we extended the high range of sounds in Study 4 (compared to Study 3), we have not yet encountered this limit on the group level; however, at least one respondent noted that \emph{``deeper sounds leaned more into the negative perceptions [...] Interestingly, though, very high pitched noises tended to the negative as well.''} For now, the consensus seems to be that higher-pitched robots are favorable, leading to descriptors such as \emph{``lively,'' ``cute,'' ``bright,'' ``curious,'' ``sweet,'' ``happy,''} and \emph{``friendly.''} The incidence of these labels was highest for Study 4, which included the highest-pitched stimuli. On the other hand, participants described lower-pitched robots as \emph{``evil,'' ``ominous,'' ``scary,''} and \emph{``as if it was working off a Darth Vader audio kit.''} So far, it appears that the pitching up of robot arm sounds leads to a sort of positive ''sci-fi'' perception of robots, in which robots seem like a friendly \emph{``alien thing''} or WALL-E-like entity.  Pitching sounds down, on the other hand, leads to the opposite. This suggests that users may associate the higher-pitched sounds with with a cute (or \textit{kawaii} in Japanese~\cite{cheok_kawaiicute_2012}) robot. 

\subsection{Design Implications}

Our results indicate two clear themes that robot designers should carefully consider: 1) \emph{quieter robot arms are less discomforting} than loud robots and 2) \emph{higher-pitched robot arms seem more pleasant, energetic, and warm, as well as less discomforting} than lower-pitched robots. After repeating a similar study design across different robot arms and use contexts, we are reasonably confident that these results would extend to additional robot arms and use cases. Follow-up investigations in person, in addition to replication across other robotic systems, would serve as a helpful final reinforcement of these findings.

For robot arm designers, we recommend selecting parts to reduce loudness emitted by the system. Particularly when the trade-off is nominal, selecting mechanical elements that reduce sound intensity will improve the comfort of users. This design principle aligns well with workplace safety standards (e.g., regulations on workplace sound levels~\cite{frazee_enabling_2020}); however, we acknowledge that it is not always desirable or feasible to eliminate robot sound altogether. For example, the emission of audible robot sound can help people safely localize robots in their environment (e.g., in~\cite{cha_effects_2018}). In such cases, we suggest keeping intensity levels close to the minimum audible level in a given environment. Furthermore, when it is impossible or impractical to fully eliminate sound, tactics like the following pitch adjustment principles might serve as another option for improving robot favorability.

In terms of pitch, reducing lower-pitched sounds in the design of robot arms seems to be most crucial. Transforming the consequential sounds of a robot to a higher-pitch can also enhance perceptions of the system. While changes in pitch intensity may inherently be linked to part selection, alterations to pitch intensity based on real-time sampling might be an option to increase positive perceptions of robots. For example, future robotic systems might include a microphone and hardware for fast audio sampling and processing.  Coupled with the additional capability to then emit this modified sound through speakers would enable the transformation of the overall robot sound profile. Using this type of pipeline, one could overlay or augment robot sounds in real time to achieve desired robot attributes.

\subsection{Strengths \& Limitations}

One key strength of this work is its rigorous evaluation of the relationship between physical sound attributes and HRI. At least within the application space of robot arms, we identified repeatable results related to modulations of loudness and pitch of robot sound. While research on robot sound has historically seemed crucial but full of mixed results, our initial results and proposed methods for research in this space offer the potential to alter the state of the field. Our proposed sound editing approaches offer an accessible path for most roboticists without the need of proprietary strategies and tools of audio engineering.

A major limitation of this work is the reliance on video-based surveys (rather than in-person interactions with robots) to evaluate consequential sound effects. We plan to conduct follow-up in-person studies to confirm that the observed trends extend to in-lab and in-the-wild interactions with robots. As with most studies, the participant population was not fully representative of all robot users. Due to our MTurk eligibility criteria, most results of this work are centered on the cultural context of the United States. We plan to conduct follow-up work with more diverse participant groups, and we welcome others to build on this work with their own stakeholder populations of interest.

\subsection{Conclusions}

In this work, we performed four online studies to investigate the effects of robot sound. This allowed us to compare more levels (i.e., eight levels of loudness or pitch per study) and attain more repeatable results than past research on robot sound. Within the space of robot arms, our results demonstrated that people perceive quieter and higher-pitched robots more positively. These pitch results surprised us somewhat; we expected that eventually a high-pitched robot would reach the point of annoyance, but even our highest-pitched stimuli were not yet perceived as such. Rather, the high-pitched sound appeared to largely make the robot seem cute or \emph{kawaii}~\cite{cheok_kawaiicute_2012}. Thus, our overall recommendation is to make robots quiet and high-pitched (or ``\emph{kawaii}-et''). 
This work can inform robot designers, researchers, and audio engineers who wish to improve perceptions of robots.

\section*{Acknowledgments}

We thank Prof. Joe Davidson for access to his UR5e robot, Kevin Green and Jeremy Dao for the Cassie robot recording, Ameer Helmi for help with figures, and Dylan Moore for study design guidance.

\bibliographystyle{IEEEtran}
\bibliography{references}

\end{document}